\documentclass[10pt,twocolumn,letterpaper]{article}

\usepackage[pagenumbers]{iccv} 

\definecolor{iccvblue}{rgb}{0.21,0.49,0.74}
\usepackage[pagebackref,breaklinks,colorlinks,allcolors=iccvblue]{hyperref}
\def\thickhline{\noalign{\hrule height0.3pt}}

\title{CARL: Causality-guided Architecture Representation Learning for an Interpretable Performance Predictor}

\author{Han Ji \quad Yuqi Feng \quad Jiahao Fan \quad Yanan Sun\\
College of Computer Science, Sichuan University\\
{\tt\small jihan@stu.scu.edu.cn, feng770623@gmail.com, \{fanjh,ysun\}@scu.edu.cn}
}

\begin{document}
\maketitle
\begin{abstract}
Performance predictors have emerged as a promising method to accelerate the evaluation stage of neural architecture search (NAS). These predictors estimate the performance of unseen architectures by learning from the correlation between a small set of trained architectures and their performance. However, most existing predictors ignore the inherent distribution shift between limited training samples and diverse test samples. Hence, they tend to learn spurious correlations as shortcuts to predictions, leading to poor generalization. To address this, we propose a \underline{C}ausality-guided \underline{A}rchitecture \underline{R}epresentation \underline{L}earning (CARL) method aiming to separate critical (causal) and redundant (non-causal) features of architectures for generalizable architecture performance prediction. Specifically, we employ a substructure extractor to split the input architecture into critical and redundant substructures in the latent space. Then, we generate multiple interventional samples by pairing critical representations with diverse redundant representations to prioritize critical features. Extensive experiments on five NAS search spaces demonstrate the state-of-the-art accuracy and superior interpretability of CARL. For instance, CARL achieves 97.67\% top-1 accuracy on CIFAR-10 using DARTS.
\end{abstract}

\section{Introduction}
\label{sec:intro}
Neural architecture search (NAS) can automatically design well-performing architectures for a specific task and even surpass manually designed solutions~\cite{elsken2019neural}. NAS typically involves three key components: the search space, the search strategy, and the performance evaluation. Conventional methods evaluate architecture performance by training from scratch~\cite{zoph2016neural,xie2017genetic,zoph2018learning}, making the evaluation stage time-consuming and resource-intensive. To mitigate this issue, a considerable number of approaches aim to reduce evaluation cost by shrinking the search space~\cite{hu2020angle,li2020improving}, sharing weights~\cite{liu2018darts,chu2020fair}, and employing zero-cost proxies~\cite{abdelfattah2021zero,mellor2021neural}. However, they still have notable drawbacks. Reducing search space diminishes the diversity of architectures~\cite{tan2019mnasnet}. Weight-sharing methods suffer from severe performance collapse~\cite{chu2020fair}, thus hardly discovering the optimal architecture in the search space~\cite{ren2021comprehensive}. Zero-cost proxies tend to have weak correlations between the predicted accuracy and the actual architecture performance~\cite{Lee2024aznas}.

\begin{figure}[t]
\centering
\includegraphics[width=\linewidth]{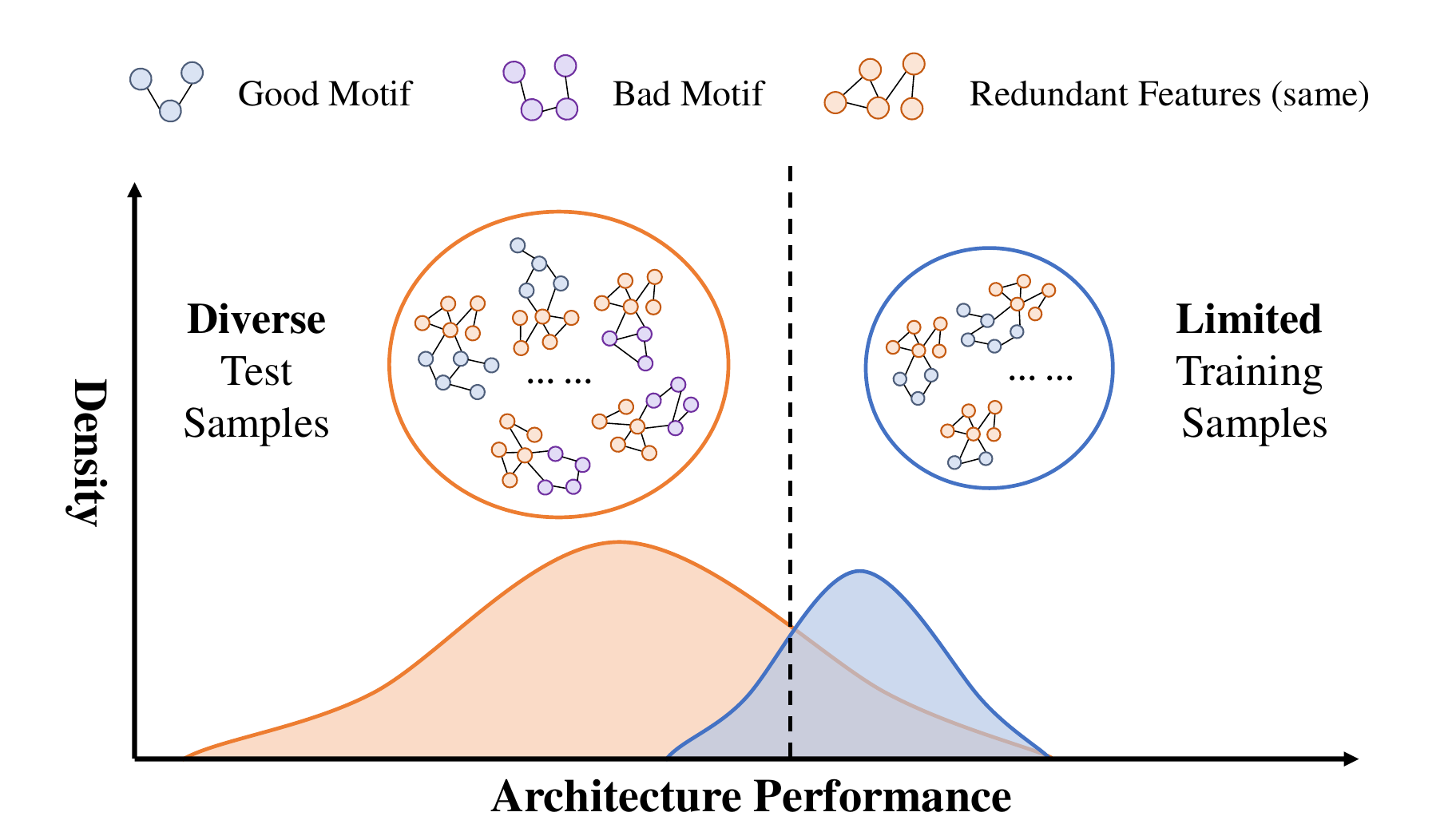}
\caption{ Illustration of the distribution shift between training and test samples for performance predictors. Good/Bad motifs lead to good/poor performance. In practice, only a small portion of samples are randomly collected to train the predictor. For architectures containing the same redundant features, there can exist a distinct performance gap between training and test samples. Hence, the trained predictors inevitably use specific redundant features as shortcuts to predictions, causing poor generalization on diverse test samples which have a much wider performance range.}
\label{fig:0}
\end{figure}

Performance predictors have achieved remarkable success in improving NAS efficiency~\cite{wen2020neural,liu2021homogeneous,wei2022npenas}, which can directly evaluate unseen architectures after learning from a small amount of training samples. Architecture representation learning has played a key role in performance evaluation. To obtain informative architecture representation, most predictors mainly focus on designing advanced models~\cite{lu2021tnasp,yi2023nar}, mimicking the information flow in architectures~\cite{ning2020generic,ning2022ta}, and tailoring self-supervised tasks~\cite{jing2022graph,zheng2024dclp}.

Recent studies~\cite{ru2020interpretable,wan2021redundancy} further reveal that only a subset of critical features, such as a motif, play a dominant role in architecture performance, while the remaining redundant features contribute marginally. Hence, some predictors~\cite{chen2021not,mills2024building} aim to improve the architecture representation by capturing such critical features. HOP~\cite{chen2021not} and AutoBuild~\cite{mills2024building} utilize an adaptive attention module and a magnitude ranked embedding space to learn the relative significance of features within architectures, respectively. Despite their success, they neglect the distribution shift between the training and test samples. The training samples are inevitably biased because they are randomly chosen from the search space and take up a very small portion. Taking the training and test samples in Figure~\ref{fig:0} as an example, specific redundant features frequently appear in well-performing architectures but have a marginal impact on performance in the training set. These biased samples can induce predictors to learn such redundant features as indicators of well-performing architectures rather than capturing the critical good motifs. Consequently, predictors are prone to overfitting redundant cues, leading to poor generalization on diverse test samples.

To boost the generalization of performance predictors, we propose a Causality-guided Architecture Representation Learning (CARL) method to effectively capture critical features of architectures and eliminate the confounding effect of redundant features under the guidance of the causal theory. Concretely, the input architectures are first encoded into representations and then split into critical and redundant representations through a substructure extractor. Then, we combine each critical feature with various redundant features to generate multiple interventional training samples. We further propose an intervention-based loss function, encouraging the predictor to primarily focus on critical features. Hence, CARL ensures a stable causal mapping from critical features to ground-truth performance. We validate the effectiveness of CARL on multiple search spaces and the experimental results exhibit its excellent performance and superior interpretability.

Our main contributions can be summarized as follows: 

(1) We pioneer to formulate the architecture performance prediction in NAS from a causal perspective, revealing how redundant features mislead performance predictors by inducing spurious correlations. 

(2) We propose CARL, a novel architecture representation learning method that discovers critical features causally correlated with performance through causal intervention to improve the generalization ability of predictors.

(3) Extensive experiments on five NAS search spaces show that CARL outperforms state-of-the-art predictors in terms of architecture performance prediction and predictor-based NAS. Besides, the ablation studies demonstrate the effectiveness of the key components of CARL.

\section{Related Works}
\label{sec:related}
\subsection{Performance Predictor for NAS}
\label{sec:2.1}
A performance predictor typically consists of an encoder that learns architecture representations and a regressor that maps these representations to architecture performance.
To improve prediction accuracy, existing predictors mainly focus on enhancing the quality of architecture representation~\cite{ning2020generic,wen2020neural,lu2021tnasp}. 
One popular approach uses advanced models, such as Graph Neural Networks (GNNs)~\cite{wen2020neural,dudziak2020brp,wei2022npenas} and Transformers~\cite{lu2021tnasp,yi2023nar}, for representation learning. PINAT~\cite{lu2023pinat}, a Transformer-based predictor, captures the spatial topology information within architectures through global attention mechanisms. Another common method ~\cite{ning2020generic,ning2021evaluating,ning2022ta,hwang2024flowerformer} models the information flow of the data processing in architectures to obtain representation. Additionally, recent works~\cite{yan2021cate,zheng2024dclp} leverage self-supervised learning to enhance architecture representations. For instance, GMAE-NAS~\cite{jing2022graph} pre-trains the predictor by masking and reconstructing operations of unlabeled architectures. However, most of these predictors neglect the distribution shift between training and test samples, causing them to rely on redundant features as shortcuts to predictions, which leads to sub-optimal architecture representations and poor generalization. To tackle this issue, we propose a causality-guided architecture representation learning method that enables the predictor to establish a causal relationship between performance and critical features, instead of overfitting to the confounding redundant features.

\subsection{Causal Intervention}
\label{sec:2.2}
Causal intervention~\cite{pearl2000models} aims to investigate the impact of variables from a causal perspective. Backdoor adjustment is the key component in causal intervention, which enables the model to pursue the causal effects by cutting out the backdoor path. Recently, a growing number of research have brought causal intervention into various computer vision tasks for model generalization. For example, VC R-CNN~\cite{wang2020visual} combines stratified confounding context objects with causal factors to enhance the generalization ability of the model in visual commonsense learning. EIIL~\cite{creager2021environment} performs the environmental intervention on data to exclude useless background information from the model inference stage. IFL~\cite{tang2022invariant} leverages attribute-invariant features to boost model performance in the generalized long-tailed classification task. The core ideas of these methods agree on disentangling causal factors and confounders to establish a stable relationship from causal factors to model performance. The model predictions are encouraged to be invariant no matter how confounders change. Inspired by these studies, this paper aims to investigate the causal relationship between the input architectures and the output predictions. This is helpful to understand the effects of different features and improve architecture representation learning by eliminating the confounding effects of redundant features.
\section{Preliminary and Analysis}
\label{sec:3.1}
\subsection{Problem Formulation}
\label{sec:problem}
The goal of a performance predictor is to estimate the performance of unseen architectures according to architecture features. Specifically, given an input architecture $x$ with ground-truth performance $y$, the predictor first utilizes an encoder to obtain the representation $z$ of $x$. After that, the predictor maps the obtained representation $z$ to the prediction value $\hat{y}$ through a regressor.

\subsection{Causal Analysis}
\label{sec:analysis}
In this part, we formulate the process of architecture performance prediction from a causal perspective. We employ the structural causal model (SCM)~\cite{pearl2009causality}, a widely-used tool in causal learning, to illustrate the causal relationship. As shown in Figure~\ref{fig:2}, the SCM of architecture performance prediction is composed of four variables, i.e., architecture data $X$, critical features $C$, redundant features $R$, and ground-truth performance $Y$. The directed edges in the SCM describe the causalities between these variables. For example, the link $C\rightarrow Y$ indicates that $C$ causes $Y$. Below, we provide a detailed explanation of these causal relationships:
\begin{figure}[t]
    \centering
    \includegraphics[width=\linewidth]{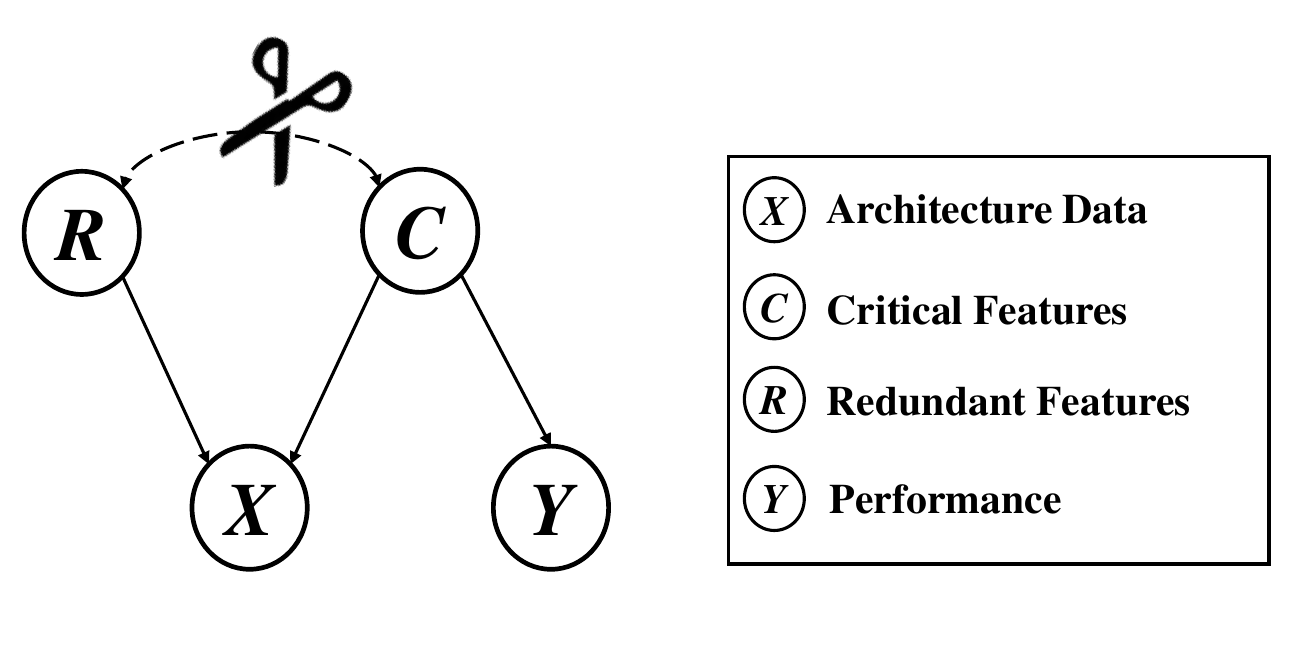}
        \caption{Causal view of architecture performance prediction.}
    \label{fig:2}
\end{figure}

\begin{figure*}[t]
\centering
\includegraphics[width=0.95\linewidth]{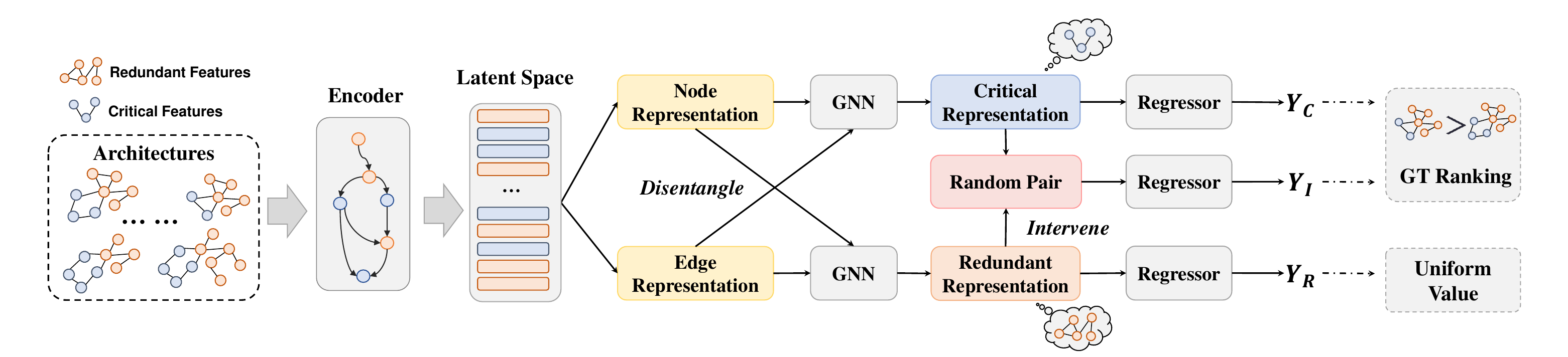}
\caption{The overall pipeline of CARL. GT denotes the ground-truth. The blue and orange regions reflect the flow of critical and redundant features of architectures within the pipeline, respectively.}
\label{fig:3}
\end{figure*}

\noindent\textbf{$\boldsymbol{R\rightarrow X \leftarrow C}$}. Architecture data $X$ contains two unobserved variables: causal features $C$ and redundant features $R$. $C$ and $R$ are disjoint and coexist in architecture $X$.

\noindent\textbf{$\boldsymbol{C\rightarrow Y}$}. This link indicates that critical features $C$ have a dominant impact on architecture performance. $C$ can be a good or bad motif containing specific operations and connections within the architecture, which explain why the architecture has good or poor performance.

\noindent\textbf{$\boldsymbol{C\dashleftarrow \dashrightarrow R}$}. This dotted link reflects the probabilistic dependencies between causal features $C$ and redundant features $R$~\cite{pearl2016causal}, which can be a direct cause ($C\rightarrow R$) or an unobserved common cause $M$ ($R\leftarrow M \rightarrow C$). Since we can not distinguish between these scenarios in practice, here we treat them equivalently. For example, specific redundant features frequently appear with good motifs in training samples. 

In the ideal scenario, the predictor should make predictions solely according to critical features $C$. However, ${C\dashleftarrow \dashrightarrow R}$ introduces spurious correlations between redundant features $R$ and performance $Y$ via the backdoor path $R\leftrightarrow C \rightarrow Y$. These spurious correlations mainly result from the sampling bias of the performance predictor. For instance, the predictor majorly selects specific redundant features with good motifs as training samples, while such redundant features are combined with both good and bad motifs in test samples. Therefore, the predictor tends to make wrong predictions for test samples based on redundant features. To eliminate the effect of redundant features $R$ on $Y$, the backdoor path between $R$ and $Y$ should be cut off as shown in Figure~\ref{fig:2}. Hence, we propose to perform causal intervention on $R$.

An intuitive intervention approach involves directly synthesizing additional training samples by combining critical and redundant features from different architectures. However, this approach is infeasible for architecture performance prediction. First, the synthesized samples may disrupt the valid computation flow of an architecture, such as architectures without an output node. Additionally, in practice, critical and redundant features are difficult to observe and identify in architecture data. To address these issues, we propose an alternative strategy that approximates the desired effect at the representation level.

\section{Methodology}
\label{sec:method}
Based on the above causal analysis, we propose a causality-guided architecture learning method \textbf{CARL} to establish a causal relationship between critical features and performance. The overall pipeline of CARL is illustrated in Figure~\ref{fig:3}. We address the challenges mentioned in Section~\ref{sec:analysis} by performing causal intervention in the latent space of architectures. First, we use an encoder to obtain the node-level and edge-level architecture representations. Once the architecture representations are obtained, we employ a subgraph extractor to disentangle the representations into critical and redundant parts, respectively. Then, we pair critical representations with redundant representations derived from other architectures to generate multiple interventional samples. Finally, we propose an intervention-based loss function to optimize the predictor. The implementation details of CARL are described as follows.

\subsection{Architecture Encoding}
We encode an architecture as a directed acyclic graph where each node indicates an operation and each edge represents a connection between operations. To be specific, we denote an input architecture by $X = (O, A)$. $A \in \{0,1\}^{D\times D}$ is an adjacent matrix and $O \in \{0,1\}^{D\times F}$ is a one-hot feature vector, where $D$ is the number of nodes and $F$ is the number of operation type. To obtain the architecture representation, we choose Graph Convolution Network (GCN)~\cite{kipf2016semi} as the encoder for its success in prior works~\cite{wen2020neural,dudziak2020brp,liu2022bridge}. For an $L$-layer GCN, the features are updated by Equation(~\ref{eq:1_5}):
\begin{equation}
    \label{eq:1_5}
    H^{l+1} = ReLU (AO^lW^l)
\end{equation}
where $H^l$ and $W_l$ represent the node representation and weight matrix of the $l$-th layer GCN, respectively. ReLU is a non-linear activation function. 

\subsection{Representation Disentanglement} 
We denote the output of the $\mathcal{L}$-th layer GCN as $H_{node}=\{H_{v_1}, H_{v_2}, \dots\}$, which represent the node-level representation of $X$. Based on this, we obtain the edge-level representation $H_{edge}=\{H_{e_{12}}, H_{e_{13}}, \dots\}$, where $H_{v_1}$ and $H_{e_{12}}$ denote the embedding of node $v_1$ and the edge from node $v_1$ to $v_2$, respectively. Each edge embedding is obtained by concatenating the embeddings of both edge vertices. For example, the embedding of the edge $e_{ij}: v_i \rightarrow v_j$ is computed by $H_{e_{ij}}=concatenate(H_{v_i}, H_{v_j})$.

After that, we utilize a substructure extractor for the predictor to learn to distinguish causal or redundant substructures by selectively highlighting nodes and edges. To achieve this, we first employ a multi-layer perceptron (MLP) to evaluate the contribution of each node to the causal and redundant substructures:
\begin{equation}
    \label{eq:2}
    \alpha_{C_i},\alpha_{R_i} = Softmax(MLP(H_{v_i})),
\end{equation}
where $\alpha_{C_i}$ and $\alpha_{R_i}$ measure the significance of node $v_i$ in generating causal and redundant substructures, respectively. Softmax is an activation function that projects the two output scores of MLP into a range of [0,1]. Note that $\alpha_{C_i}+\alpha_{R_i}=1$. Similarly, at the edge level, another MLP is applied to compute the $\alpha_{C_{jk}}$ and $\alpha_{R_{jk}}$ for each edge $e_{jk}$, ensuring that $\alpha_{C_{jk}}+\alpha_{R_{jk}}=1$. 

Then, we can calculate node-level critical mask matrix $M_O^C=\{\alpha_{C_1},\alpha_{C_2},\dots\}$ and redundant mask matrix $M_O^R= \{\alpha_{R_1},\alpha_{R_2},\dots\}$. At the edge level, we can also obtain such causal and redundant mask matrix $M_A^C$ and $M_A^R$. Based on these mask matrices, we can split architecture $X$ into a causal substructure $X_C = (O\odot M_O^C, A\odot M_A^C)$ and a 
redundant counterpart $X_R = (O\odot M_O^R, A\odot M_A^R)$. Then, we use GCNs $g_c$ and $g_r$ to embed these substructures into critical representations $Z_C = g_c(X_C)$ and redundant representations $Z_R = g_r(X_R)$. 

\begin{table*}[t]
    \centering
    \tiny
    \renewcommand{\arraystretch}{1.09}
    \caption{Comparison of different predictors on NAS-Bench-101 and NAS-Bench-201. The training portion indicates the portion of architectures used to train a predictor. \textbf{Bold} indicates the best. $^{*}$: implemented by ourselves using the open sources.}
    \resizebox{1.86\columnwidth}{!}{
    \begin{tabular}{l|ccc|ccc}
    \thickhline
            \textbf{Search Space}    & \multicolumn{3}{c|}{\textbf{NAS-Bench-101}} & \multicolumn{3}{c}{\textbf{NAS-Bench-201}} \\ 
            \hline
           \vspace{0.5pt}\textbf{Training Portion}  &\textbf{0.02\%} &\textbf{0.04\%} &\textbf{0.1\%} &\textbf{0.5\%} &\textbf{1\%} &\textbf{3\%} \\
           \hline
         NAO~\cite{luo2018neural}   & 0.501 & 0.566 & 0.666 & 0.467 & 0.493  & 0.470 \\
         NP~\cite{wen2020neural}   & 0.391 & 0.545 & 0.679 & 0.343 & 0.413  & 0.584 \\
         Arch2Vec~\cite{yan2020does}  & 0.435 & 0.511 & 0.547  & 0.542 & 0.573 & 0.601 \\
         GraphTrans~\cite{Wu2021GraphTrans} & 0.330 & 0.472 & 0.602 & 0.409 & 0.550 & 0.594 \\
         Graphormer~\cite{ying2021transformers} & 0.564 & 0.580 & 0.611 & 0.505 & 0.630 & 0.680 \\
         TNASP~\cite{lu2021tnasp} & 0.600 & 0.669 & 0.705 & 0.539 & 0.589 & 0.640 \\
         NAR-Former$^{*}$~\cite{yi2023nar} & 0.627 & 0.651 & 0.765 & 0.590 & 0.652 & 0.720 \\
         PINAT~\cite{lu2023pinat} & 0.679 & 0.715 & 0.772 & 0.549 & 0.631 & 0.706 \\
         \hline
         CARL (ours)   & \textbf{0.683} & \textbf{0.723} & \textbf{0.774} & \textbf{0.656}  & \textbf{0.697} & \textbf{0.755} \\
    \thickhline
    \end{tabular}
    }
    \label{table:ktau res}
\end{table*}

 Once architecture representations are separated, we employ a MLP-based regressor to map the representations to the corresponding performance. Based on the analysis in Section~\ref{sec:analysis}, the predictor is encouraged to make predictions solely based on critical representation $Z_C$ because critical features dominantly contribute to the architecture performance. As suggested in previous works~\cite{xu2021renas,ning2022ta}, we use hinge ranking loss to optimize the learning of critical features because predicting relative ranking between architectures is important. The loss function $\mathcal{L}_C$ for learning critical features is formulated as Equation~\eqref{eq:4}:
\begin{equation}
\label{eq:4}
    \mathcal{L}_C = \sum_{i=1}^{N-1}\sum_{j=i+1}^{N} max(0,m-(\hat{y^i_c}-\hat{y^j_c})*{sign}(y^i-y^j)),
\end{equation}
where $N$ is the number of training data and $m$ is the margin. For each architecture pair $(x_i,x_j)$ in the training set, the loss will be 0 when they are ranked correctly and differed by a specific margin.

Since redundant features make negligible contributions to performance, the predictor should output a uniform prediction value for all redundant features to avoid their impact on performance. Hence, we use the mean square error as the loss function for redundant features. The loss function $\mathcal{L}_R$ for disregarding redundant features is computed by Equation~\eqref{eq:5}:
\begin{equation}
\label{eq:5}
   \mathcal{L}_R = \frac{1}{N} \sum_{i=1}^{N}(\hat{y_r^i}-\bar{y})^{2},
\end{equation}
where $\bar{y}$ denotes the mean accuracy of training samples. 

\subsection{Interventional Samples Generation}
Up to now, we have disentangled architecture representations into causal and redundant parts.
However, the probabilistic dependencies between critical and redundant features still create spurious correlations between redundant features and performance. Hence, we propose to generate interventional samples in the latent space. To be specific, we pair each critical representation ${Z_C}$ with a random redundant representation $Z_R'$ from other architectures in the mini-batch to yield the interventional representation $Z_I$. In this way, critical and redundant parts within the generated interventional sample have fewer statistical correlations. 

Hence, the intervention prediction $Y_{I}$ should be consistent with the ground-truth performance of the critical part, no matter where the redundant representation $Z_R$ comes from. The loss function $\mathcal{L}_I$ for learning on interventional samples is calculated by Equation~\eqref{eq:6}:
\begin{equation}
\label{eq:6}
   \mathcal{L}_I = \sum_{i=1}^{N-1}\sum_{j=i+1}^{N} max(0,m-(\hat{y^i_I}-\hat{y^j_I})*{sign}(y^i-y^j)),
\end{equation}
In summary, the intervention-based loss function for CARL is derived from the three key components discussed earlier, as shown in Equation~\eqref{eq:7}:
\begin{equation}
\label{eq:7}
     \mathcal{L} = \mathcal{L}_C + \lambda_1 \mathcal{L}_R + \lambda_2 \mathcal{L}_{I},
\end{equation}
where $\lambda_1$ and $\lambda_2$ are hyperparameters that control the relative importance of representation disentanglement and intervention, respectively.

\section{Experiments}
\label{sec:experiments}
\subsection{Experimental Settings}
\label{sec:5.1}
We evaluate the effectiveness of CARL from two aspects: the ranking performance and the performance of predictor-based NAS. The experiments are conducted on five search spaces, i.e., NAS-Bench-101~\cite{ying2019bench}, NAS-Bench-201~\cite{dong2019bench}, TransNAS-Bench-101~\cite{duan2021transnas}, NAS-Bench-NLP~\cite{klyuchnikov2022bench}, and DARTS~\cite{liu2018darts}. We adopt Kendall’s Tau~\cite{sen1968estimates} (Ktau) as the evaluation metric for ranking performance for its widespread use in prior works~\cite{yi2023nar,lu2023pinat,zheng2024dclp}. Meanwhile, we follow the convention~\cite{wen2020neural,lu2023pinat} and train the predictor with the validation accuracy of architectures and evaluate the predictor with the test accuracy. In the ranking experiments, we use the whole search space as the test set for a more consistent comparison between different methods.

\subsection{Ranking Experiments}
\label{sec:5.2}
 We compare the ranking ability of CARL with other performance predictors for six data splits on NAS-Bench-101 and NAS-Bench-201. As shown in Table~\ref{table:ktau res}, CARL takes the lead among all the strong competitors across six data splits. On NAS-Bench-101, CARL achieves higher Ktau with only 0.02\% training data than TNASP~\cite{lu2021tnasp} and NAR~Former~\cite{yi2023nar} which use 0.04\% training data. This reflects that CARL can still learn informative architecture representations that generalize well on the diverse dataset with fewer training samples. Since evaluating the ground-truth performance of architectures is extremely time-consuming, we focus more on the results of a small training portion. On NAS-Bench-201, CARL outperforms PINAT~\cite{lu2023pa} and NAR-Former by 0.066 and 0.107 with 0.5\% training samples. The superior results of CARL can be attributed to its stable causal relationship between critical features and performance.

\subsection{NAS Experiments}
\label{sec:5.3}
To evaluate CARL in the NAS settings, we conduct experiments on image classification tasks within three search spaces (i.e., NAS-Bench-101, NAS-Bench-201, and DARTS) and other tasks within another two search spaces (i.e., TransNAS-Bench-101 and DARTS). 

\subsubsection{Experiments on Image Classification Tasks}
 We employ the popular predictor-guided evolutionary search framework~\cite{real2019regularized} on NAS-Bench-101 and NAS-Bench-201. As for DARTS~\cite{liu2018darts}, we follow the convention~\cite{wen2020neural,liu2022bridge} and train the predictor with 100 architectures that are trained for 50 epochs on CIFAR-10. 100,000 architectures are then randomly sampled from the search space and predicted by the predictor. The architecture with the highest predicted value is selected and retrained to acquire the final test accuracy. The settings of retraining are consistent with DARTS~\cite{liu2018darts}. The architecture is searched on CIFAR-10 and then directly transferred to ImageNet.

\begin{table}[t]
 \centering
\caption{Searching results on NAS-Bench-101. They are averaged for ten runs. $^{\dagger}$: reported by DCLP~\cite{zheng2024dclp}.}
 \resizebox{0.95\columnwidth}{!}{
        \begin{tabular}{lccc}
        \toprule
        \textbf{Method}    & \textbf{Test Acc.(\%)} & \textbf{Query} & \textbf{ Strategy} \\
        \midrule
        RS~\cite{li2020random} & 93.64 & 150      & Random                \\
        REA~\cite{real2019regularized}           & 93.80 & 150        &Evolution\\  
        \midrule
        HAAP$^{\dagger}$~\cite{liu2021homogeneous}          & 93.69 & 300    &Evolution            \\
        BANANAS~\cite{white2021bananas}      & 94.08 & 150      &BO                   \\
        HOP$^{\dagger}$~\cite{chen2021not} &94.09   & 300       & Random                \\
        WeakNAS~\cite{wu2021stronger} &94.10 & 150 & Random  \\
        CATE~\cite{yan2021cate}          & 94.12 & 150       & Reinforce                \\  
        FlowerFormer~\cite{hwang2024flowerformer}       & 94.14 & 150       & Evolution                 \\  
        GMAE-NAS~\cite{jing2022graph}      & 94.14 & 150       & Evolution               \\
        NPENAS~\cite{wei2022npenas}        & 94.15 & 150        & Evolution                 \\
        DCLP~\cite{zheng2024dclp}         & 94.17 & 300       & Evolution              \\
        \midrule
        CARL (Average)  & \textbf{94.17} & \textbf{150}   & Evolution                     \\
        CARL (Best)  & \textbf{94.23} & \textbf{150}   & Evolution                     \\
        \bottomrule
\end{tabular}
}
\label{table:search 101}
\end{table}

\paragraph{Results on NAS-Bench-101.} The search results for different NAS methods are presented in Table~\ref{table:search 101}. Compared with other predictor-based methods, CARL helps discover the most competitive architecture on CIFAR-10 with the same or a smaller query budget. In particular, CARL yields the highest accuracy of 94.17\% with 2$\times$ fewer queried architectures than DCLP~\cite{zheng2024dclp}. This improvement is not trivial compared with prior works. 

\begin{table}[ht]
\centering
\renewcommand{\arraystretch}{1.05}
 \caption{Searching results on NAS-Bench-201. They are averaged for ten runs. $^{\dagger}$: reported by DCLP~\cite{zheng2024dclp}. The query budget of DCLP is accumulated by 50 architectures during training and 30 architectures for initialization.}
 \resizebox{0.99\columnwidth}{!}{
        \begin{tabular}{lcccc}
        \toprule
        \textbf{Method}   & \textbf{Query} & \textbf{CIFAR-10} & \textbf{CIFAR-100}  & \textbf{ImageNet-16-120} \\
        \midrule
         RS~\cite{bergstra2012random}  &$>$500     & $93.70\pm0.36$ & $71.04\pm1.07$   & $44.57\pm1.25$        \\
         REA~\cite{real2019regularized}  &$>$500     & $93.92\pm0.30$ & $71.84\pm0.99$  & $45.54\pm1.03$         \\
         \midrule
          HAAP$^{\dagger}$~\cite{liu2021homogeneous}  &150     & $93.75\pm0.17$  & $71.08\pm0.19$  & $45.22\pm0.26$       \\
         GMAE-NAS$^{\dagger}$~\cite{jing2022graph}      &150     & $94.03\pm0.21$  & $72.56\pm0.16$  & $46.09\pm0.27$         \\
         HOP$^{\dagger}$~\cite{chen2021not}      & 150  & $94.10\pm0.12$  & $72.64\pm0.11$  & $46.29 \pm0.19$                 \\
         ReNAS~\cite{xu2021renas} & 100 & $94.34\pm0.03$  & $73.50\pm0.30$  & $46.54 \pm0.03$               \\
         DCLP~\cite{zheng2024dclp} & 80 & $94.34\pm0.03$  & $73.50\pm0.30$  & $46.54 \pm0.03$               \\
         \midrule
         CARL (Average)          &\textbf{80}  & $\textbf{94.37}\pm\textbf{0.00}$  & $\textbf{73.51}\pm\textbf{0.00}$  & $\textbf{46.91}\pm\textbf{0.32}$                   \\
         CARL (Best)         & 80 & 94.37 & 73.51 & 47.31             \\
         \midrule
         Optimal & - & 94.37  & 73.51  & 47.31  \\
         \bottomrule
\end{tabular}}
    \label{table:search 201}
\end{table}

\paragraph{Results on NAS-Bench-201.} We compare CARL with other methods in Table~\ref{table:search 201} and find that CARL can achieve state-of-the-art performance across three datasets with merely 80 queried architectures. Specifically, CARL can find the optimal architectures on CIFAR-10 and CIFAR-100~\cite{Krizhevsky2009LearningML} datasets. Besides, CARL improves the searching efficiency with 6$\times$ fewer queried architectures compared with conventional NAS methods like random search~\cite{bergstra2012random} and regularized evolution aging~\cite{real2019regularized}. Among all the predictor-based methods, CARL performs best with fewer or equivalent queried architectures, which indicates its capability of guiding NAS to explore the search space efficiently. 

\begin{table}[ht]
 \caption{Results on ImageNet using DARTS search space.}
 \resizebox{0.95\columnwidth}{!}{
  \begin{tabular}{cccc}
        \toprule
        \textbf{Method} &\textbf{Top-1/5 Acc. (\%)}  & \textbf{\# Params (M)}  &\textbf{Cost (GPU Days)} \\
        \midrule
        
        DARTS~\cite{liu2018darts}  & 73.3 / 91.3 & \textbf{4.7} & 4 \\
        GDAS~\cite{dong2019searching}  & 74.0 / 92.5  & 5.3 & 0.21\\
        PC-DARTS~\cite{xu2019pc} & 74.9 / 92.2  & 5.3  & \textbf{0.1} \\
        P-DARTS~\cite{chen2019progressive}  & 75.6 / 92.6  & 4.9 & 0.3 \\
        FairDARTS~\cite{chu2020fair}  & 73.7 / 91.7  & 4.8 & 0.4 \\ 
        Shapley-NAS~\cite{xiao2022shapley} & 75.7 / 92.7  & 5.1 & 0.3  \\
        \midrule
        NAO~\cite{luo2018neural}  & 74.3 / 91.8  & 11.35 & 200 \\
        PRE-NAS~\cite{peng2022pre}   & 76.0 / 92.6  & 6.2 & 0.6 \\
        TNASP-B~\cite{lu2021tnasp} & 75.5 / 92.5  & 5.1 & 0.3  \\
        PINAT-T~\cite{lu2023pinat} & 75.1 / 92.5  & 5.2 & 0.3  \\
        \midrule
        
        CARL (ours) & \textbf{76.1}/ \textbf{92.8}  & 5.3  & 0.25 \\
        \bottomrule
        \end{tabular}
    }
\label{table:darts image}
\end{table}
\vspace{-1em}

\begin{table}[ht]
 \centering
\caption{Results on CIFAR-10 using DARTS search space.}
 \resizebox{0.95\columnwidth}{!}{
    \begin{tabular}{ccccc}
        \toprule
        \textbf{Method}   & \textbf{ Acc. (\%)} & \textbf{\# Params (M)} & \textbf{Cost (GPU Days)} \\  
        \midrule
        DARTS~\cite{liu2018darts}             & 97.24 & \textbf{3.3}  & 4       \\
         PC-DARTS~\cite{xu2019pc}      & 97.43 & 3.6 & \textbf{0.1} \\
         GM~\cite{hu2021generalizing}  & 97.60 & 3.7 & 2.7 \\
         DrNAS~\cite{chen2021drnas}   & 97.54 & 4.1 & 0.6\\
         PA\&DA~\cite{lu2023pa}   & 97.66 & 3.9 & 0.4 \\
        \midrule
        NAO~\cite{luo2018neural}          & 97.50 & 10.6 & 200       \\
        Arch2Vec-BO~\cite{yan2020does}  & 97.52 & 3.6 & 100 \\
        NAS-BOWL~\cite{ru2020interpretable}          & 97.50 & 3.7  & 3        \\
        BANANAS~\cite{white2021bananas}          & 97.43 & 3.6   & 11.8        \\
        TNASP~\cite{lu2021tnasp}          & 97.48 & 3.6  & 0.3        \\
        NPENAS-NP~\cite{wei2022npenas}  & 97.56 & 3.5  & 1.8        \\
        PINAT~\cite{lu2023pinat} & 97.58 & 3.6 & 0.3 \\
        NAR-Former~\cite{yi2023nar}       & 97.52  & 3.8  & 0.24        \\
        \midrule
        CARL (ours)            & \textbf{97.67}  & 3.7  & 0.25       \\
        \bottomrule              
    \end{tabular}
    }
\label{table:darts cf10}
\end{table}

\begin{table*}[t!]
    \centering
     \caption{Searching results on TransNAS-Bench-101 Micro. They are averaged for ten runs. $^{\dagger}$: reported by Arch-Graph~\cite{huang2022arch}.}
       \resizebox{1.95\columnwidth}{!}{
    \begin{tabular}{ccccccccc}
    \toprule
    \rule{0pt}{8pt}
   Tasks & Cls.O. & Cls.S. & Auto. & Normal & Sem. Seg. & Room. & Jigsaw & \multirow{2}{*}{ Avg. Rank} \\
    \cline{1-8}
    \rule{0pt}{8pt}
  Metric  & Acc.$^{\uparrow}$ & Acc.$^{\uparrow}$ & $\operatorname{SSIM}^{\uparrow}$ & $\operatorname{SSIM}^{\uparrow}$ & $\mathrm{mIoU}^{\uparrow}$ & L2 loss$\downarrow$ & Acc.$^{\uparrow}$ & \\
    \midrule  
  BONAS$^{\dagger}$~\cite{shi2020bridging} & 45.50 & 54.46 & 56.73 & 57.46 & 25.32 & 61.10 & 94.81 & 34.31 \\
   Arch-Graph~\cite{huang2022arch} & 45.48 & 54.70 & 56.52 & 57.53 & 25.71 & 61.05 & 94.66 & 22.15 \\
  WeakNAS$^{\dagger}$~\cite{wu2021stronger} & 45.66 & 54.72 & 56.77 & 57.21 & \textbf{25.90} & \textbf{60.31} & 94.63 & 20.03 \\
   \midrule 
 CARL (ours)  & \textbf{45.69} & \textbf{54.78} & \textbf{57.01} & \textbf{57.77} & 25.80 & 60.92 & \textbf{94.84} & \textbf{12.29} \\
    \midrule
 Global Best & 46.32 & 54.94 & 57.72 & 59.62 & 26.27 & 59.38 & 95.37 & 1 \\
\bottomrule
\end{tabular} 
}
    \label{table:search tb101}
\end{table*}

\begin{table*}[t]
    \centering
    \caption{Searching results on TransNAS-Bench-101 Macro. They are averaged for ten runs. $^{\dagger}$: reported by Arch-Graph~\cite{huang2022arch}.}
       \resizebox{1.95\columnwidth}{!}{
    \begin{tabular}{ccccccccc}
    \toprule
    \rule{0pt}{8pt}
   Tasks & Cls.O. & Cls.S. & Auto. & Normal & Sem. Seg. & Room. & Jigsaw & \multirow{2}{*}{ Avg. Rank} \\
    \cline{1-8}
    \rule{0pt}{8pt}
  Metric  & Acc.$^{\uparrow}$ & Acc.$^{\uparrow}$ & $\operatorname{SSIM}^{\uparrow}$ & $\operatorname{SSIM}^{\uparrow}$ & $\mathrm{mIoU}^{\uparrow}$ & L2 loss$\downarrow$ & Acc.$^{\uparrow}$ & \\
    \midrule  
  BONAS$^{\dagger}$~\cite{shi2020bridging} & 46.85 & 56.47 & \textbf{74.45} & 61.62 & 28.82 & 59.39 & 96.76 & 33.37 \\
   Arch-Graph~\cite{huang2022arch} & 47.35 & 56.77 & 71.32 & \textbf{62.78} & 29.09 & 58.05 & 96.70 & 12.68 \\
  WeakNAS$^{\dagger}$~\cite{wu2021stronger} & 47.40 & 56.88 & 72.54 & 62.37 & 29.18 & 57.86 & 96.86 & 10.49 \\
   \midrule 
 CARL (ours) & \textbf{47.45} & \textbf{56.92} & 74.12 & 62.44 & \textbf{29.22} & \textbf{57.68} & \textbf{96.89} & \textbf{5.29} \\
    \midrule
 Global Best & 47.96 & 57.48 & 76.88 & 64.35 & 29.66 & 56.28 & 97.02 & 1 \\
\bottomrule
\end{tabular} 
}
    \label{table:search macro} 
\end{table*}

\paragraph{Results on DARTS ImageNet.}
Table~\ref{table:darts image} demonstrates that the architecture searched by CARL can achieve a top-1 accuracy of 76.1\% and a top-5 accuracy of 92.8\% on ImageNet, outperforming previous state-of-the-art one-shot and predictor-based NAS methods. Note that CARL requires a low search cost of only 0.25 GPU days, which is the least among the predictor-based NAS methods. 

\paragraph{Results on DARTS CIFAR-10.}
As shown in Table~\ref{table:darts cf10}, we observe that the architecture discovered by CARL can achieve a superior test accuracy of 97.67$\%$, which is 0.09$\%$ higher than the previous state-of-the-art predictor-based method: PINAT~\cite{lu2023pinat}. Although some one-shot NAS methods like GM~\cite{hu2021generalizing} and PA$\&$DA~\cite{lu2023pa} also yield competitive results, CARL exhibits superior performance with a lower search cost.

\subsubsection{Experiments on Other Tasks}
 We employ the predictor-guided evolutionary search~\cite{real2019regularized} on TransNAS-Bench-101 Micro and the predictor-guided random search ~\cite{bergstra2012random} on TransNAS-Bench-101 Macro and NAS-Bench-NLP. TransNAS-Bench-101 includes seven challenging computer vision tasks, and NAS-Bench-NLP provides architecture performance on the Natural Language Processing (NLP) task. Note that TransNAS-Bench-101 Macro is a skeleton-based search space, instead of the common cell-based ones.   

\paragraph{Results on TransNAS-Bench-101.} The query budget is set to 50 architectures for both spaces. It can be observed from Table~\ref{table:search tb101} that CARL helps discover architectures with promising results on different tasks in the micro space. Although WeakNAS~\cite{wu2021stronger} takes the lead in some specific scenarios, CARL accomplishes the best average ranking of 12.29 across all seven tasks, demonstrating its superior generalization in the challenging NAS task.

To search for architectures in the macro space, a skeleton-based search space, we opt for the same architecture encoding in Arch-Graph~\cite{huang2022arch} for CARL. Table~\ref{table:search macro} demonstrates that CARL yields the best average rank of 5.29, which is 2$\times$ higher than WeakNAS~\cite{wu2021stronger}. This indicates the excellent scalability of CARL beyond cell-based search spaces.

\begin{table}[h]
    \centering
    \renewcommand{\arraystretch}{1.1}
    \caption{Searching results on NAS-Bench-NLP. The results are averaged for ten runs. Lower log PPL indicates better performance.}
    \resizebox{0.98\columnwidth}{!}{
    \begin{tabular}{l|ccccc}
        \toprule

        \textbf{Method}   &\textbf{BANANAS~\cite{white2021bananas}} &\textbf{NPENAS~\cite{wei2022npenas}} &\textbf{NASBOT~\cite{kandasamy2018neural}}  &\textbf{CARL (ours)} \\
        \hline
         Log PPL         & 4.595 & 4.587 & 4.579 & \textbf{4.572}  \\
         Strategy         & BO   & Evolution & BO & Random \\
        \bottomrule
    \end{tabular}
    }
    \label{table:search nlp}
\end{table}
\vspace{-1em}

\paragraph{Results on NAS-Bench-NLP.} The query budget is set to 100 architectures for all methods. The results in Table~\ref{table:search nlp} illustrate that CARL yields a low log perplexity (PPL) of 4.572, outperforming other competitive predictors like NPENAS~\cite{wei2022npenas}. This suggests that CARL can also generalize well beyond computer vision tasks. 

\subsection{Ablation Study}
\label{sec:5.5}
To validate the effects of the different components in CARL, we conduct ablation studies on NAS-Bench-201. We explore the impacts of the encoder, the effectiveness of the substructure extractor, and the effects of hyperparameters in the loss function.

\begin{figure}[t]
     \begin{minipage}[b]{0.48\columnwidth}
        \centering
        \includegraphics[width=0.98\columnwidth]{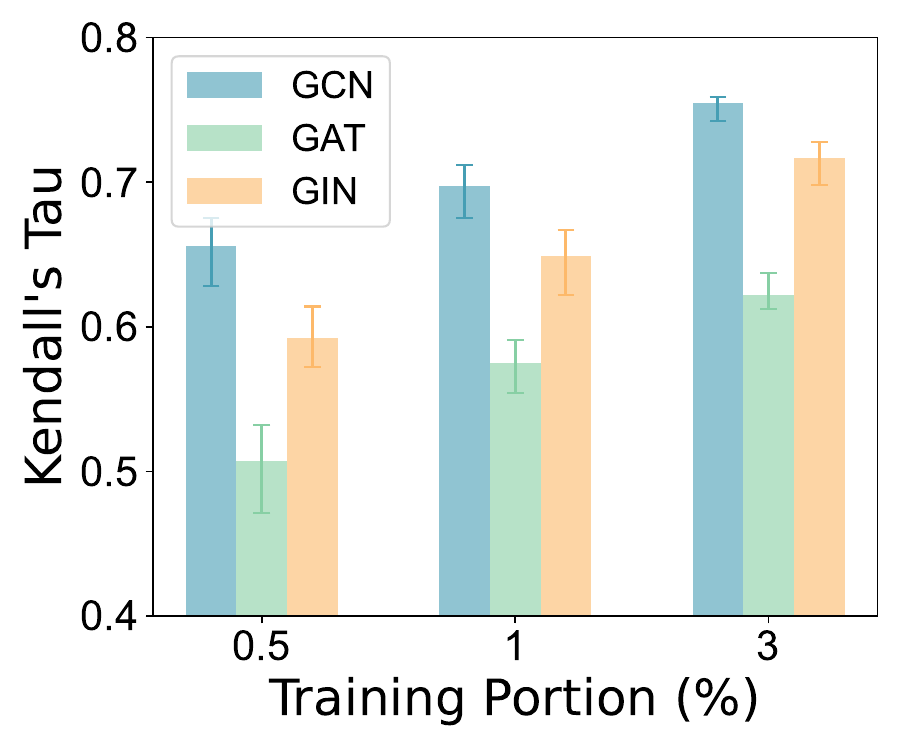}
        \centerline{(a) Encoder}
    \end{minipage}
        \begin{minipage}[b]{0.48\columnwidth}
        \centering
        \includegraphics[width=0.98\columnwidth]{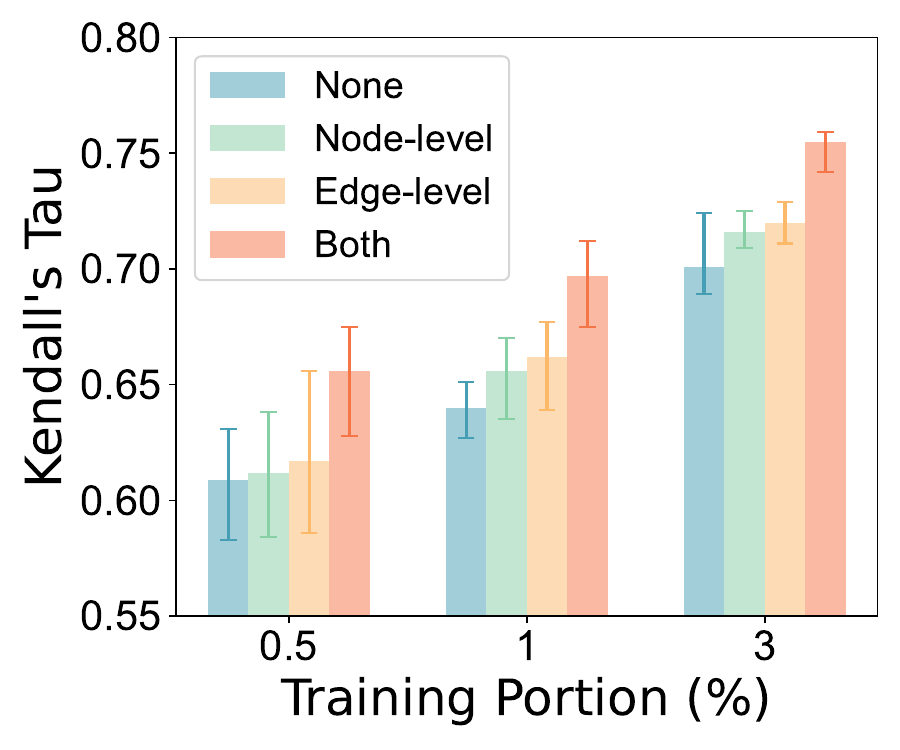}
        \centerline{(b) Substructure Extractor}
    \end{minipage}
    \caption{Ablation results of encoder and substructure extractor.}
    \label{fig:4}
\end{figure}

 \begin{table}[t]
    \caption{Ablation Results of $\lambda_1$ and $\lambda_2$ on NAS-Bench-201 with 0.5\% training samples.}
    \centering
    \tiny
     \renewcommand{\arraystretch}{1.02}
       \resizebox{0.95\columnwidth}{!}{
    \begin{tabular}{c|ccccc}
\thickhline
    {$\lambda_1 \& \lambda_2$}  & 0 & 0.2 & 0.5 & 0.7 & 1 \\ 
   \hline
  0 & 0.601 & 0.604 & 0.618 & 0.611 & 0.608  \\
  0.2 & 0.610 & 0.628 & 0.634 & 0.635 & 0.612  \\
  0.5 & 0.636 & 0.647 & \textbf{0.656} & 0.633 & 0.629 \\
  0.7 & 0.616 & 0.629 & 0.651 & 0.626 & 0.619  \\
  1 & 0.606 & 0.616 & 0.635 & 0.617 & 0.617 \\
\thickhline
\end{tabular} 
}
\label{tab:lambda}
\end{table}

\paragraph{Effect of Encoders.}
To explore the impacts of encoders on CARL, we compare the results of three popular encoders: GCN~\cite{kipf2016semi}, GIN~\cite{xu2018powerful}, and GAT~\cite{vaswani2017attention}. As shown in Figure~\ref{fig:4}(a), we identify that GCN consistently outperforms other encoders by a large margin. Meanwhile, GAT and GIN perform poorly when the training portion is low. Hence, GCN is a more suitable choice for CARL.

\paragraph{Necessity of Node-level and Edge-level Disentanglement.}
In CARL, a substructure extractor is employed to disentangle architecture representation on the node and edge level. We further examine the effects of node-level and edge-level disentanglement alone. 
For instance, the importance score of each node is divided into $\{0.5,0.5\}$ for critical and redundant features in the exploration of edge-level disentanglement. 
 We can observe from Figure~\ref{fig:4}(b) that using node-level and edge-level disentanglement together yields the highest Ktau. Meanwhile, edge-level disentanglement slightly outperforms node-level one because the edge representation derives from the representation of two connected nodes and contains more attribute and topology information. These results indicate that both node-level and edge-level disentanglement play a significant role in CARL.

 \paragraph{Impact of Disentanglement and Intervention Intensity.} In Eq.~\ref{eq:7}, we use hyperparameters $\lambda_1$ and $\lambda_2$ to control the intensity of disentanglement and intervention. To investigate the sensitivity of these hyperparameters, we perform a study on NAS-Bench-201. As shown in Table~\ref{tab:lambda}, we observe a performance drop when $\lambda_1$ and $\lambda_2$ are either too small or too large. If $\lambda_1$ and $\lambda_2$ are set too small, the predictor can not correctly suppress the redundant features. If they are set too large, the predictor fails to capture the crtical features. 

\subsection{Visualization Results}
\label{sec:5.6}
\begin{figure}[t!]
        \centering
        \includegraphics[width=0.98\columnwidth]{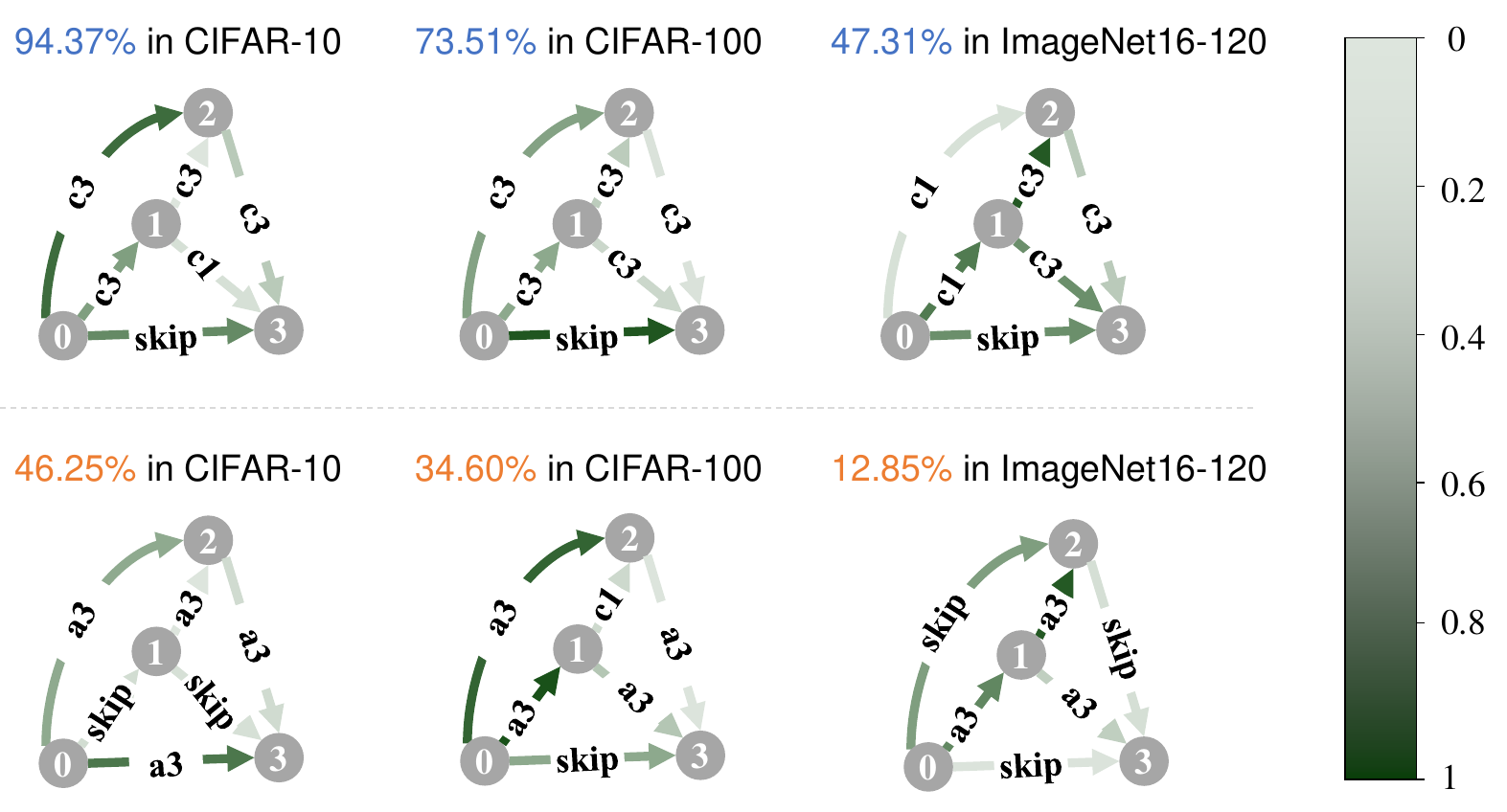}
    \caption{Visualization results for architectures on NAS-Bench-201 where edge types represent operations and edge direction represent the computation flow. `Skip' denotes skip connect, `a3' denotes avg-pool 3$\times$3, `c3' denotes conv 3$\times$3, and `c1' denotes conv 1$\times$1. Node 0 is the input node and node 3 is the output node. An edge with a darker color indicates a higher importance score for this operation. (\textit{Top}) well-performing architectures. (\textit{Bottom}) poorly-performing architectures.}
    \label{fig:5}
\end{figure}
\vspace{-0.5em}

To illustrate the relative significance of architecture features to ground-truth performance, we visualize the important score for each operation through CARL in Figure~\ref{fig:5}. We apply CARL to both well-performing and poorly-performing architectures in three datasets on NAS-Bench-201. We identify that the features connected with node 0 (input node) tend to have higher important scores, which indicates a larger significance to architecture performance. Besides, critical features with frequent convolution operations lead to a promising architecture, while excessive avg-pool operations in critical features cause poor performance. The critical features found by CARL are also consistent with the motifs discovered in previous studies~\cite{ru2020interpretable,wan2021redundancy}, which provide interpretability for predictions. These results suggest that CARL can effectively capture critical features that dominantly determine architecture performance. 

\section{Conclusion}
\label{sec:conclusion}
In this paper, we investigate the process of architecture performance prediction in NAS from a causal perspective for the first time. We identify that the spurious correlations between redundant features and performance resulting from the distribution shift can cause poor generalization of performance predictors. To address this issue, we propose CARL to disentangle architecture features and further perform the causal intervention to capture critical features effectively. Experimental results on various NAS search spaces demonstrate the superior performance and interpretability of CARL. In the future, we will study more advanced causal intervention approaches to enhance CARL.

{
    \small
    \bibliographystyle{ieeenat_fullname}
    \bibliography{main}
}

\end{document}